\documentclass[journal]{IEEEtran}

\ifCLASSINFOpdf
\else
\fi

\usepackage{tikz}
\usepackage{amssymb}
\usepackage{amsmath}
\usepackage{setspace}
\usetikzlibrary{snakes}

\newcommand{\vx}{{\bf x}}

\newcommand{\vw}{{\bf w}}
\newcommand{\vv}{{\bf v}}

\newcommand{\matSigma}{{\bf \Sigma}}

\usepackage{hyperref}
\definecolor{darkblue}{rgb}{0,0,.5}
\hypersetup{pdftex=true, colorlinks=true, breaklinks=true, linkcolor=darkblue, menucolor=darkblue, pagecolor=darkblue, urlcolor=darkblue}

\begin{document}
\title{Transferring Subspaces Between Subjects in Brain-Computer Interfacing}
\author{Wojciech Samek,~\IEEEmembership{Student Member,~IEEE,}
        Frank C.~Meinecke
        and~Klaus-Robert M\"{u}ller,~\IEEEmembership{Member,~IEEE,}
\thanks{W. Samek, F.~C.~Meinecke and K-R. M\"{u}ller are with Berlin Institute of Berlin, Franklinstr. 28 / 29, 10587 Berlin, Germany. E-Mail: wojciech.samek@tu-berlin.de, frank.meinecke@tu-berlin.de, klaus-robert.mueller@tu-berlin.de.}
\thanks{K-R. M\"{u}ller is with the Department of Brain and Cognitive Engineering, Korea University, Anam-dong, Seongbuk-gu, Seoul 136-713, Korea}
\thanks{Copyright (c) 2013 IEEE. Personal use of this material is permitted. However, permission to use this material for any other purposes must be obtained from the IEEE by sending an email to pubs-permissions@ieee.org.}
\thanks{W.~Samek, F.~C.~Meinecke and K-R.~M\"uller, Transferring Subspaces Between Subjects in Brain-Computer Interfacing, {\it IEEE Transactions on Biomedical Engineering}, 2013. \url{http://dx.doi.org/10.1109/TBME.2013.2253608}}}
\markboth{Samek et al. $-$ Transferring Subspaces Between Subjects in BCI}{Samek et al. $-$ Transferring Subspaces Between Subjects in BCI}
\maketitle

\begin{abstract}
Compensating changes between a subjects' training and testing session in Brain Computer Interfacing (BCI) is challenging but of great importance for a robust BCI operation. We show that such changes are very similar between subjects, thus can be reliably estimated using data from other users and utilized to construct an invariant feature space. This novel approach to learning from other subjects aims to reduce the adverse effects of common non-stationarities, but does not transfer discriminative information. This is an important conceptual difference to standard multi-subject methods that e.g.\ improve the covariance matrix estimation by shrinking it towards the average of other users or construct a global feature space.
These methods do not reduces the shift between training and test data and may produce poor results when subjects have very different signal characteristics.
In this paper we compare our approach to two state-of-the-art multi-subject methods on toy data and two data sets of EEG recordings from subjects performing motor imagery. We show that it can not only achieve a significant increase in performance, but also that the extracted change patterns allow for a neurophysiologically meaningful interpretation.
\end{abstract}


\IEEEpeerreviewmaketitle
%%%%%%%%%%%%%%%%%%%%%%%%%%%%%%%%%%%%%%%%%%%%%%%%%%%%%%%%%%%%%%%%%%%%%%%%%%%%%%%%

\section{Introduction}
\IEEEPARstart{I}ncorporating data from other subjects (or sessions) into the learning process has gained much attention in the Brain-Computer Interfacing (BCI) community \cite{KraTanBlaMue08, Dev11, Ala10} as it reduces calibration times and allows to construct subject-independent spatial filters and/or classifiers. One popular approach \cite{Lotte2010, kan09} is to regularize the covariance matrix towards the average covariance matrix of other subjects in order to improve its estimation quality. This kind of regularization is especially promising in small-sample settings. Another very recent approach to transfer learning in BCI \cite{Dev11} formulates the Common Spatial Patterns (CSP) computation as a multi-subject optimization problem, thus incorporates information from other subjects in order to construct a common feature space. It must be noted that both methods rely on very strong assumptions, namely a common underlying data generating process and similarity between the discriminative subspaces, respectively. However, due to the non-stationary nature of EEG and large variations between subjects these assumptions are hardly satisfied. This makes learning a common representation or classification model very challenging, e.g.\ when two subjects have different signal characteristics, these methods may even deteriorate performance as the spatial filters or classifier will be regularized in the ``wrong'' direction. A careful subject selection or weighting is therefore essential for a successful application.

In this paper we propose a diametrically opposite approach, namely instead of learning the task-relevant part from others, we transfer information about non-stationarities in the data. Our method is especially promising when significant changes are present in the data e.g.\ induced by differences in experimental conditions between sessions. Its underlying assumption is that these principal non-stationarities are similar between subjects, thus can be transferred, and have an adverse effect on classification performance, thus removing them is favourable.
Unlike the methods presented before our approach reduces the shift between training and test data and does not assume similarity between discriminative subspaces. Note that we define the discriminative subspace as the subspace spanned by the CSP filters.
One important advantage of our method is the fact that the negative impact on performance is limited when subjects have very different signal characteristics. This is because the spatial filters are not regularized ``towards'' a low dimensional subspace, but ``away'' from one. In other words under the assumption that the true discriminative subspace is small\footnote{This assumption is reasonable as the feature space extracted by CSP usually does not contain more than a few dimensions.} compared to the data space, it is very unlikely that we remove a significant amount of discriminative information with our method. On the other hand when regularizing towards a small discriminative subspace we effectively disregard much larger amount of information (orthogonal complement of this subspace), thus if subjects have very different signal characteristics we may lose relevant information. Consequently, the importance of subject clustering or subject selection is largely reduced in our method.

One scenario where transfer of information about non-stationarities is especially useful is an experiment with differences in the stimulus presentation or feedback mode between sessions. For instance if a visual cue is presented in the test phase, but is lacking when calibrating the system then we may expect increased occipital activity in the test data due to additional visual processing. This increase in activity should be taken into account when computing the spatial filters as otherwise it may lead to non-stationary features. Since this increase is relatively stable between subjects, we can learn its patterns from other users and use them to extract invariant features.

In summary, regularization towards discriminative subspaces of other users and utilization of knowledge about prominent changes are two complementary tasks which have different assumptions and scenarios of application. The regularization approach has already been successfully applied in BCI studies and is especially promising when data is scarce and the subject similarity is high. The transfer of non-stationary information on the other hand is novel and is especially useful when common non-stationarities can be expected from the experiment. 

This paper is organized as follows. In the next section we present related work and review two state-of-the-art methods for between-subject transfer in BCI. In Section III we describe the underlying assumptions of our approach and introduce the algorithm. In Section IV we present and analyse results from toy experiments and experiments on real EEG recordings from two different data sets containing prominent non-stationarities between training and test session. We conclude in Section V with a discussion.

%%%%%%%%%%%%%%%%%%%%%%%%%%%%%%%%%%%%%%%%%%%%%%%%%%%%%%%%%%%%%%%%%%%%%%%%%%%%%%%%%%%
\section{Related Work}
Reliable classification under covariate shift, i.e.\ in situations where the data distribution changes between training and testing phase, is a topic of increasing popularity in many application domains of machine learning \cite{Can09, SugKaw2011}. In particular it is of interest in the field of Brain-Computer Interfacing as the measured brain signals are highly non-stationary \cite{SheKraBlaRaoMue06, SugKraMue07, reuderink2011robust}. There are basically two strategies to tackle the problem of changing signal properties, namely adaptation of the features or the classifier and extraction of robust representations that are less affected by variations of the underlying brain processes. The approaches presented in this work all belong to the second category, thus we limit the literature review to that. 

One of the most popular feature extraction methods in BCI is Common Spatial Patterns (CSP) \cite{Blankertz08optimizingspatial, Ramoser98optimalspatial, LemmBlaDicMue11} as it is well suited to discriminate between different mental states induced by motor imagery. A spatial filter $\vw$ computed with CSP maximizes the variance of band-pass filtered EEG signals in one condition while minimizing it in the other condition. Since variance of a band-pass filtered signal is equal to band power, CSP enhances the differences in band power between two conditions.
CSP is prone to overfitting and does not ensure stationarity of the feature, thus many different variants robustifying the original algorithm have been proposed \cite{LotGua11, SamJNE12, Arv13}. The idea of an invariant feature space was proposed in \cite{Blankertz08invariantcommon} and was adapted in \cite{SamJNE12} where the authors introduce a stationary version of CSP to trade-off stationarity and discriminativity of the extracted features. 
The stationary CSP method penalizes filters that lead to non-stationary features, thus ensures stability over time and consequently better classification. Since this method is computed on training data and does not incorporate data from other subjects, it is not able to capture changes occurring in the transition between training and testing stage. A different strategy to ensure stationary of the features was proposed in \cite{BueMeiKirMue09, Sam12}. The authors propose to remove the non-stationary subspace from data in a preprocessing step prior to feature computation, however, also here neither the shift between sessions is considered nor does the method incorporate data from other subjects.

Several CSP extensions utilizing information from other subjects have been proposed in the context of zero-training BCI and small-sample setting.
For instance a very recently proposed method \cite{Dev11} learns a spatial filter for a new subject based on its own data and that of other users. Another recent work \cite{Lotte2010} regularizes the Common Spatial Patterns (CSP) and Linear Discriminant Analysis (LDA) algorithms based on data from a subset of automatically selected subjects. A method that aims at zero training for Brain-Computer Interfacing by utilizing knowledge from the same subject collected in previous sessions was proposed in \cite{KraTanBlaMue08, Kra08, fazli2009subject}. The authors of \cite{Ala10} train a classifier that is able to learn from multiple subjects by multi-task learning. The method proposed in \cite{kan09} uses the similarity between subjects measured by Kullback-Leibler divergence as weight for improving the covariance estimation by shrinkage.

In the following we describe two CSP variants that incorporate data from other subjects in more detail.

The method proposed by Lotte and Guan \cite{Lotte2010} regularizes the estimated covariance matrix towards the average covariance matrix of other subjects. This kind of regularization may largely improve the estimation quality of the high dimensional covariance matrix if data is scarce. The estimation for subject $i^*$ can be written as
\begin{equation}
\tilde{\matSigma}_{i^*,c} = (1 - \lambda)\matSigma_{i^*,c} + \lambda\frac{1}{n-1}\sum_{i=1}^{n-1}\matSigma_{i,c},
\end{equation}
where $\matSigma_{i^*,c}$ is the covariance matrix of class $c$ for the subject of interest, $\matSigma_{i,c}$ are the covariance matrices of the other $i = 1 \ldots n,\ i \not= i^*$ subjects and $\lambda \in [0\ 1]$ is a regularization parameter controlling the amount of information incorporated from other users. This method is based on a very restrictive assumption, namely the similarity between covariance matrices of different subjects. The authors in \cite{Lotte2010} recognized that this assumption is often violated due to large inter-subject variability, thus they proposed a sequential algorithm for subject selection. In the following we will refer to this approach as covariance-based CSP (covCSP).

The method proposed by Devlaminck et al.\ \cite{Dev11} assumes a similarity between spatial filters extracted from different subjects. The goal of this CSP variant is to construct a more global feature spaces by decomposing the spatial filter $\vw_i$ for each subject $i$ into a global $\vw_0$ and subject specific part $\vv_i$
\begin{equation}
\vw_i = \vw_0 + \vv_i,
\end{equation}
and applying a single optimization framework to learn both types of filters
\begin{equation}
\max_{\vw_0, \vv_i} \sum_{i=1}^n\frac{\vw^T_i\matSigma_{i,c}\vw_i}{\vw^T_i(\matSigma_{i,1} + \matSigma_{i,2})\vw_i + \lambda_1||\vw_0||^2 + \lambda_2||\vv_i||^2}.
\end{equation}
The parameters $\lambda_1$ and $\lambda_2$ trade-off between the global or specific part of the filter. For a high value of $\lambda_1$ and a low value of $\lambda_2$ the vector $\vw_0$ is forced to zero and a specific filter is constructed. The opposite case forces the vector $\vv_i$ to zero and more global filters are computed. Furthermore, one can also perform regularization by choosing both $\lambda_1$ and $\lambda_2$ high. The optimization is performed by Newton's method and conjugate constraints\footnote{The $i$th spatial filter $\vw_i$ is conjugate to the spatial filters $\vw_k$ with $k = 1 \ldots i-1$ with respect to $\matSigma_{i,c}$, i.e.\ $\vw_i^T \matSigma_{i,c} \vw_k = 0$} are added when extracting multiple spatial filters.
Note that also here the assumption of similarity between spatial filters is very restrictive and a single objective function makes the optimization problem more difficult as it can not be formulated as a generalized eigenvalue problem.
The authors of \cite{Dev11} propose a cluster-based approach to tackle the problem of inter-subject variability.
In the following this method will be referred to as multi-task CSP (mtCSP).

%%%%%%%%%%%%%%%%%%%%%%%%%%%%%%%%%%%%%%%%%%%%%%%%%%%%%%%%%%%%%%%%%%%%%%%%%%%%%%%%%%%
\section{Transferring Non-Stationarities}
In this section we introduce a novel way of using transfer learning in Brain-Computer Interfacing.
We present a method that transfers non-stationary information between subjects, thus effectively bridges the gap between training and test data.
Note that we do not claim that our method is the first one to tackle the problem of non-stationarity in BCI, there are of course other methods like stationary CSP \cite{SamJNE12}, Kullback-Leibler CSP \cite{Arv13} or adaptation methods \cite{VidSanMueBla10, VidSanMueBla11}, however, we are not aware of any multi-subject method that tackles the non-stationarity problem. Since the main focus of this work is to investigate and compare different ways of utilizing information from other subjects and not to study the relations between within-session and between-session non-stationarities, we do not compare against those approaches.

\subsection{Stationary Subspace CSP}
The goal of the stationary subspace CSP (ssCSP) method is to remove the subspace that contains the principal non-stationary directions common to most subjects prior to CSP computation. The algorithm is summarized in Table \ref{tab:algo}. 

\begin{table}
\centering
\caption{Description of our algorithm. The non-stationary subspace is computed from other subjects $i$ in order to achieve invariance for user $i^*$.}
\footnotesize
\begin{tabular}{ll}
\hline
\hline\\
\texttt{(1)} & \texttt{For each subject $i = 1 \ldots n,\ i \not= i^*$ compute}\\ & \texttt{the eigenvectors $\vv_i^{(1)} \ldots \vv_i^{(d)}$ of $\matSigma_{i}^{train} - \matSigma_{i}^{test}$.} \\\\
\texttt{(2)} & \texttt{For each subject $i$ select the $l$ eigenvectors}\\ & \texttt{with largest absolute eigenvalues.} \\\\
\texttt{(3)} & \texttt{Aggregate the vectors of all subjects}\\
& \texttt{into a matrix $P$.} \\\\
\texttt{(4)} & \texttt{Apply PCA to $P$ in order to extract the $\nu$}\\ & \texttt{most common non-stationary directions $P_{\nu}$.}\\\\
\texttt{(5)} & \texttt{Make $i^*$s spatial filters invariant to changes}\\ & \texttt{by forcing them to lie in the orthogonal}\\ & \texttt{complement of the subspace spanned by $P_{\nu}$.}\\\\
\hline
\hline
\end{tabular}
\label{tab:algo}
\end{table}

In the following we briefly describe how to extract invariant features for subject $i^*$ by utilizing data from other users.
In the first step of the method prominent directions of change are extracted from other subjects $i = 1 \ldots n,\ i \not= i^*$. For that an eigendecomposition of the difference of the training and test covariance matrix $\matSigma_{i}^{train} - \matSigma_{i}^{test}$ is computed. Note that the $l$ eigenvectors $\vv_i^{(1)}, \vv_i^{(2)} \ldots \vv_i^{(l)}$  with largest absolute eigenvalues $|\mathrm{d}_i^{(1)}|, |\mathrm{d}_i^{(2)}| \ldots |\mathrm{d}_i^{(l)}|$ capture most of the changes occurring between training and test. The parameter $l$ can be a fixed value or chosen adaptively for each subject e.g.\ by setting a threshold on the power spectrum of the eigendecomposition.
Aggregating the eigenvectors obtained from different subjects gives a matrix $P~=~\begin{bmatrix} \vv_1^{(1)} \ldots \vv_n^{(l)} \end{bmatrix}$ whose columns are the basis of the subspace of common non-stationarties $\mathcal{S}_P = \mathrm{span}(P)$.
The dimensionality of this subspace $\cal{S}_P$ can be reduced by applying Principal Component Analysis (PCA) to matrix $P$. 
This step is important as the dimensionality of $\cal{S}_P$ grows linearly with the size of $P$, i.e.\ with the number of subjects. 
By application of PCA we extract the subspace of dimensionality $\nu \leq dim(P)$ containing the most relevant information about non-stationarities. 
We denote the projection matrix to this low-dimensional subspace as $P_{\nu}$.
Note that PCA must be applied without mean subtraction as the column vectors of $P$ are directional vectors without a common zero point. 
In order to construct invariant features for subject $i^*$ we regularize the CSP filters towards the orthogonal complement of $\mathcal{S}_{P_{\nu}}$ that is defined as $\mathcal{S}_{P_{\nu}^\bot}~=~\left\{x\in \mathbb{R}^D : \langle x, y \rangle = 0 \mbox{ for all } y\in \mathcal{S}_{P_{\nu}} \right\}$.
This can be achieved by adding the penalty matrix $\Delta = \lambda P_{\nu}P_{\nu}^T$ to the denominator of the CSP object function (as done in \cite{Blankertz08optimizingspatial, SamJNE12}). From this perspective our method can be regarded as a variant of the stationary CSP algorithm with a penalty matrix that has been computed from data of other subjects and has reduced rank $\nu$. Since we aim to completely remove the non-stationary directions from the data, we set $\lambda = 10^5$.

Our approach requires setting two parameters $l$ and $\nu$. The first parameter controls the number of non-stationary directions extracted per subject. This parameter can have a fixed value for all subjects or be subject dependent, e.g.\ by defining a threshold on the amount of changes one wants to capture. The second parameter sets the dimensionality of the non-stationary subspace that is removed.
Note that the parameters can not be determined by cross-validation on the subject of interest as the goal of our method is to reduce the shift between training and test data and this does not necessary correlate with a performance increase on the training data. One approach to determine the parameters is to cross-validate the classification performance in a leave-one-subject-out manner on the other subjects.

\subsection{General Considerations}
\begin{figure}[h]
\begin{center}
\includegraphics[width=160px]{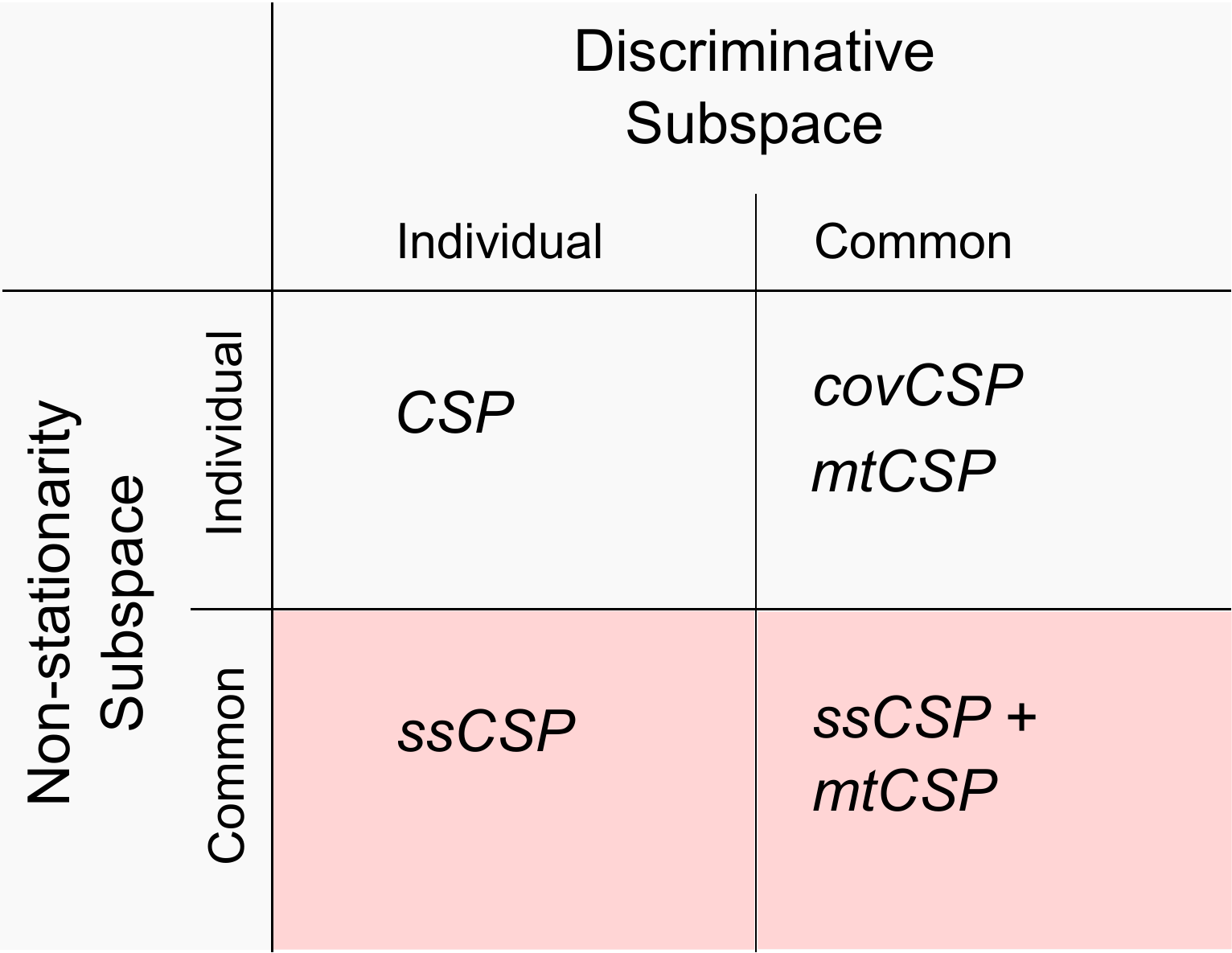}
\end{center}
\caption{Overview of the two application domains of transfer learning in BCI. If all subjects have very different discriminative and non-stationary subspaces then transfer learning is not possible, thus CSP is the method of choice. Multi-subject methods like covCSP and mtCSP are applicable if common discriminative subspaces exist. The ssCSP method is designed to remove principal changes from data, thus it assumes common non-stationary subspaces. If both the discriminative and non-stationary subspaces are similar between subjects, then a subsequent application of ssCSP and mtCSP (or covCSP) will give best results.}
\label{fig:overview}
\end{figure}

There are two types of information that can be transferred between subjects, namely discriminative and non-stationary information.
Note that both transfer types have different application scenarios e.g.\ discriminative information is important in small-sample settings as it may improve the estimation quality of the spatial filters or classifier, whereas non-stationary information is valuable when common experimental-related changes are present in the data. Figure \ref{fig:overview} illustrates the application domains of the multi-subject methods used in this work. 

If there are no common discriminative and non-stationary subspaces in the data, then transfer learning is not applicable, thus CSP is the method of choice. If on the other hand the most discriminative or non-stationary directions are similar between subjects, then the multi-subject methods described in this paper may perform much better than CSP. Finally, if both types of information can be transferred between users, then a combination of the multi-subject methods gives best results.

In order to chose the best method one needs to assess the similarity between the subjects or their discriminative and non-stationary subspaces.
This is not an easy task and is often not possible e.g.\ the directions of change cannot be estimated when test data is not available.
Furthermore it is common to perform subject selection or clustering prior to multi-subject learning in order to ensure a high level of similarity between users. However, this also requires that the subject similarity can be reliably estimated and that a large number of other subjects is available.

All three transfer learning approaches presented in this paper have regularization parameters controlling the amount of information transferred between subjects. A bad choice of these parameters may negatively affect performance, especially if subject similarity is low. Please note that the amount of information transferred in the ssCSP case is limited by the maximal dimensionality of the non-stationary subspace that is removed from the data\footnote{Since we are only interested in removing the most common changes, the maximal size of the non-stationary subspace should not exceed a fraction of the data dimensionality.}, whereas in the case of covCSP and mtCSP it is not limited, i.e.\ the classification may be completely based on data from other subjects. This is an important advantage of our multi-subject method as this limitation avoids a significant performance decrease when subject similarity is low.

\begin{figure}[h]
\begin{center}
\includegraphics[width=230px]{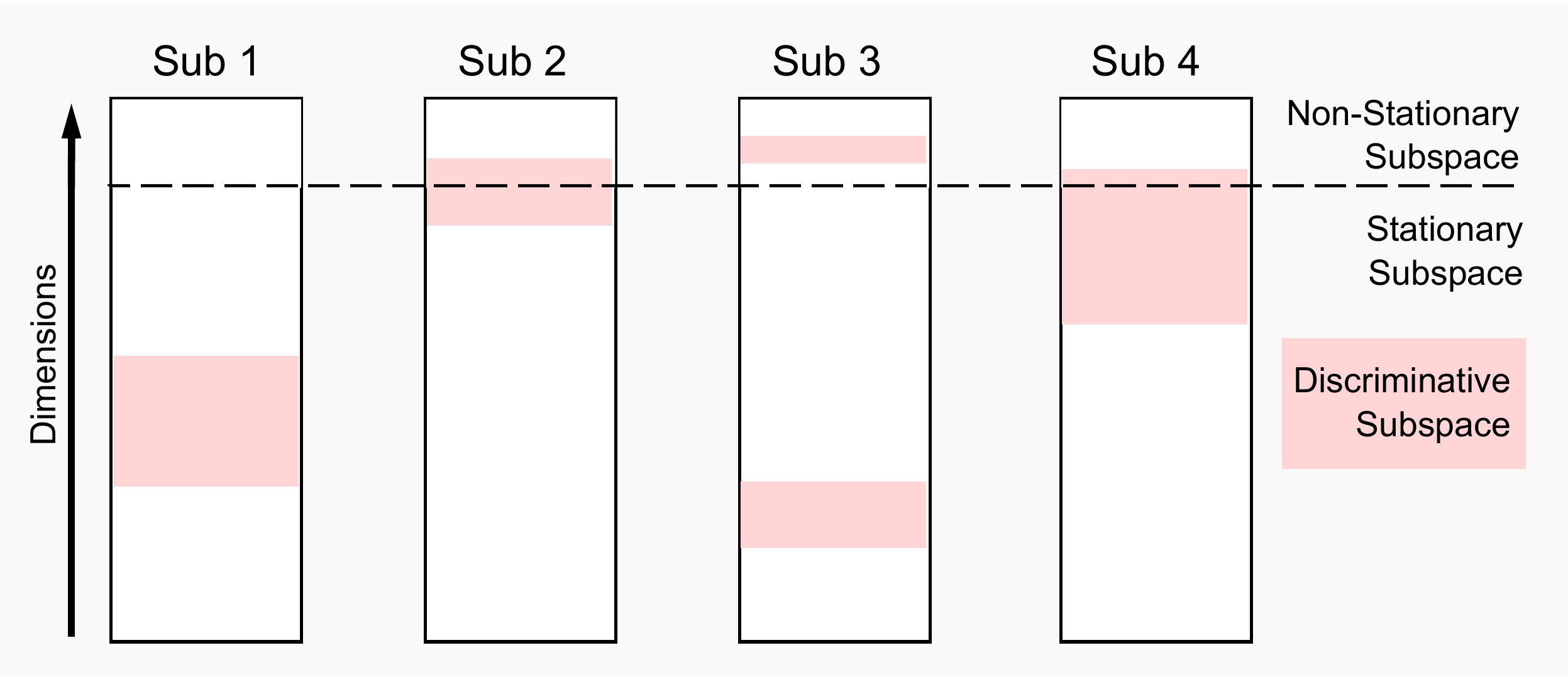}
\end{center}
\caption{An example where transferring non-stationarities between subjects is more promising than utilizing the discriminative part. The discriminative subspaces vary between subjects, whereas the non-stationary subspaces stay the same. Both subspaces are relatively small compared to the dimensionality of the data.}
\label{fig:assumptions}
\end{figure}

An example where transferring non-stationarities between subjects is more promising than utilizing the discriminative part is illustrated in Fig.~\ref{fig:assumptions}. This figure shows four artificial subjects with varying discriminative subspaces, but common directions of change. In Section IV Fig.~\ref{fig:angles} we will see that the real EEG recordings used in this paper have exactly these properties. Note that most multi-subject methods for BCI assume similarity between discriminative subspaces, thus may provide suboptimal results in such a setting. We discuss this point in the toy example in next section. One can also see from the figure that both the discriminative and non-stationary subspaces are relatively small compared to the dimensionality of the data. This is a reasonable assumption as few CSP directions usually suffice to capture the relevant information and although a larger part of the data may show non-stationary behaviour only few changes can be explained by differences between sessions. Note that we are not assuming that discriminative and non-stationary subspaces are disjoint, in contrast we explicitly aim to extract a feature space that represents the real BCI related activity and ignores discriminativity that is induced by a particular experimental setting, e.g.\ involuntarily eye movements may produce discriminative EEG patterns when using visual stimuli. Since this activity is not induced by motor imagery but is an artefact of the experimental setting, its patterns become meaningless and can harm performance when switching to a different mode of stimulus presentation. Therefore removing discriminative activity that is non-stationary makes perfectly sense when aiming for robust classification.

%%%%%%%%%%%%%%%%%%%%%%%%%%%%%%%%%%%%%%%%%%%%%%%%%%%%%%%%%%%%%%%%%%%%%%%%%%%%%%%%%%%
\section{Experimental Evaluation}
\subsection{Toy Experiment}
In this subsection we study the stability of the three multi-subject methods under increasing dissimilarity between subjects. In other words we evaluate the impact on classification performance when moving from transferring relevant information to transferring meaningless information.
The data set consists of artificially generated training and test recordings of five subjects. In order to separately study the effect of dissimilarity of the discriminative subspace and the non-stationary subspace, we generate the data as sum of two independent mixtures.
In more detail, data $\vx$ is generated as sum of a stationary noise-signal term and a non-stationary noise term
\begin{eqnarray}
  {\mathbf x}(t) & =  & \underbrace{A \, \begin{bmatrix} {\mathbf s^{dis}}(t) \\ {\mathbf s^{ndis}}(t) \end{bmatrix}}_{noise-signal\ term} + \underbrace{B \, \begin{bmatrix} {\mathbf s^{stat}}(t) \\ {\mathbf s^{nstat}}(t) \end{bmatrix}}_{noise\ term}.
\label{eq:mixing_model}
\end{eqnarray}
Note that we call the first mixture the ``noise-signal term'' as it contains contributions from sources that are relevant for a particular BCI task (signal) as well as contributions from non-relevant sources (noise). The second mixture is called ``noise term'' as its sources are not important for classification. Thus the toy data is generated by a mixture model with non-stationary noise. The matrices A and B are random rotation matrices mixing the (non-)discriminative and (non-)stationary sources and the sources are normally-distributed (with zero mean), mutually independent and independent in time. In order to approximate the properties of real data we restrict the discriminative and non-stationary subspaces to be low-dimensional. 

The following parameters are used for the experiments.
The discriminative subspace is spanned by 6 sources $s^{dis}$ with variance 0.8 in one condition and 0.1 in the other one and the non-discriminative subspace consists of 74 sources $s^{ndis}$ with fixed variance of 0.1. The 75 stationary sources $s^{stat}$ have variance 1 in both the training and test data set, whereas the variance of the 5 non-stationary sources $s^{nstat}$ is 1 in the training data set and 3 in the test data set. For each artificial subject we generate 100 trials per condition, each consisting of 100 data points, for both the training and the test set.
As in the real experiments described later in this section we extract three CSP filters per class and use log-variance features and a LDA classifier. We determine the parameters for the multi-subject methods by cross-validating classification performance in a leave-one-subject-out manner on the other users. The following experiments were performed on this toy data set using 100 repetitions.

\begin{figure*}[!ht]
\begin{center}
\includegraphics[width=480px]{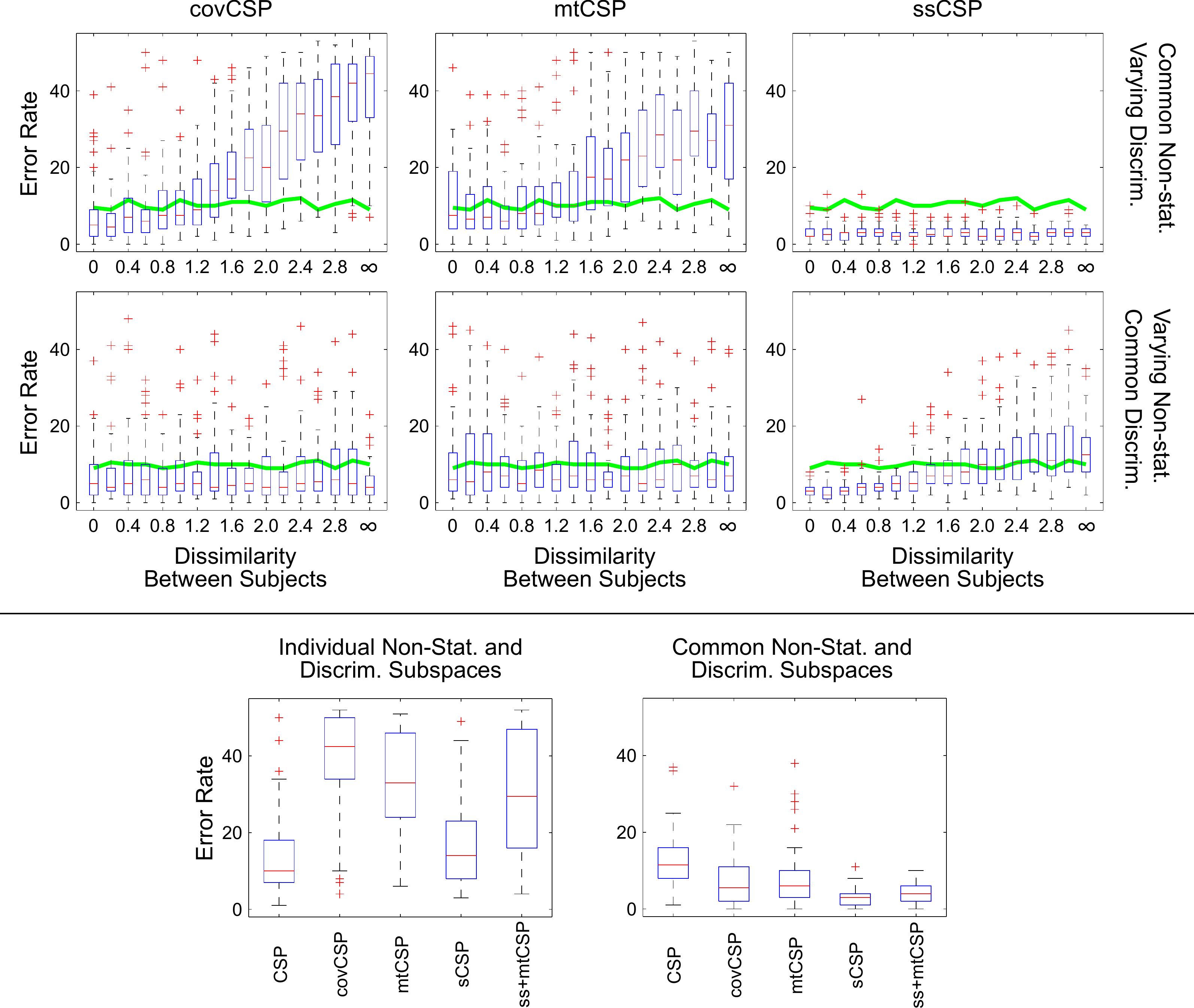}
\end{center}
\caption{Results of the three multi-subject methods on toy data. The upper row shows the case when discriminative subspaces become more and more dissimilar but the non-stationarities stay the same for all subjects. One can see that covcsp and mtcsp improve classification performance when subjects are similar, but when the difference between them becomes larger then the information transferred becomes more and more meaningless, thus error rates increase almost to chance level. The ssCSP method improves classification accuracy as it removes non-stationarities and is not affected by differences in the discriminative subspaces. The middle row shows results for the opposite case, namely constant discriminative subspaces but different non-stationary directions. The ssCSP method improves classification accuracy when the information transferred is meaningful, but does not lead to a significant increase in error rates when this is not the case. This effect is due to the asymmetry of regularizing towards and away from a small subspace. The bottom row shows the performance of all methods in the extreme case when both subspaces are either different or common between subjects.}
\label{fig:toy}
\end{figure*}

In the first experiment we fix matrix B for all subjects, but increase the distance between the mixing matrix A = $e^{\rm{M}}$ of subject 1 and the mixing matrices of the other subjects by adding an increasing amount of randomness while making sure that it still remains a rotation matrix\footnote{Matrix A is constructed as a matrix exponent of a random antisymmetric matrix M, i.e.\ A = $e^{\rm{M}}$. This ensures that A is a rotation matrix, i.e.\ AA$^\top$ = I as $A^\top = (e^{\rm{M}})^\top = e^{-\rm{M}} = A^{-1}$.}. By adding a random matrix $\Xi$ to M we obtain M$_2$ = M + $\eta$ $\Xi$. The new rotation matrix A$_2$ can be computed as A$_2$ = $e^{\frac{1}{2}(\rm{M_2} - \rm{M_2}')}$. The weight $\eta$ controls the distance between A and A$_2$. In other words we simulate the case of increasing dissimilarity between discriminative subspaces of subject 1 and the other artificial users. The results for the three multi-subject methods are summarized in the top row of Fig.~\ref{fig:toy}. Each boxplot shows the distribution of classification error rates of subject 1 for increasing dissimilarity values $\eta$. Furthermore the median CSP error rate is plotted as green curve. We see from the figure that methods that transfer discriminative information between subjects, namely covcsp and mtcsp, significantly decrease error rates when the dissimilarity between the mixing matrices A of subject 1 and the others is low. However, if the information that is transferred becomes more and more random the methods become arbitrarily bad. The stationary subspace CSP method is not affected by increased dissimilarity of the mixing matrices A as it does not transfer discriminative information. It is able to improve classification performance as the non-stationary subspace remains the same for all subjects (matrix B is constant).

In the second experiment we simulate the opposite case, namely we fix A and increase the dissimilarity of B between subject 1 and the others.
The middle row of Fig.~\ref{fig:toy} shows the results for this case. We can observe a stable improvement of the methods covcsp and mtcsp because the discriminative subspaces are the same for all subjects irrespectively of B. The figure shows an improved performance (decrease in error rates) for the ssCSP method when the dissimilarity between the non-stationary subspaces is low and a performance drop when it is high. However, the important point here is that in contrast to the discriminativity transfer in the last experiment the performance loss is minimal, actually the performance goes back to CSP level. This increased robustness can be explained with a lower risk of losing important information when regularizing the solution away from a small subspace. Although the transferred non-stationary information becomes more and more meaningless when distance between the mixing matrices B increases, classification accuracy does not decrease on average since only few directions are removed from data. Note that this asymmetric behaviour of covCSP, mtCSP and ssCSP highly depends on the size of the discriminative and non-stationary subspaces, the selection of regularization parameters and of course if subject (pre)selection is used or not.

In the final experiment we let both matrices A and B be either different or the same between subject 1 and the other users (bottom row of Fig.~\ref{fig:toy}). In the first case multi-subject methods have no advantage over CSP as there is no meaningful information to be transferred. On the contrary, the methods transferring discriminative information may even lose performance as the solution is regularized towards a non-informative subspace. In the other case when both subspaces stay constant over subjects we observe a significance performance gain of all multi-subject methods.
Since the non-stationarity problem is more severe than the estimation problem, we obtain best results for both the ssCSP method and the combination of ssCSP and mtCSP (denoted as ss+mtCSP), i.e.\ the application of mtCSP in the stationary subspace determined by ssCSP.

\subsection{Data Set}
Two different data sets are used for the real-data experiment. The first one consists of two calibration (i.e.\ without feedback) recordings from five healthy participants. The volunteers performed motor imagery of two limbs, specifically ``left hand'' and ``foot''. The cues indicating the stimulus were presented either visually (with an arrow appearing in the center of the screen) or auditory (a voice announcing the task to be performed), resulting in two different datasets for each user. In this experiment, the training data set was the calibration with visual stimuli and the test data set the calibration with auditory stimuli. A time segment located from 750ms to 3500ms after the cue instructing the subject to perform motor imagery is extracted from each trial and the signal is band-pass filtered in 8-30 Hz using a 5-th order Butterworth filter. Both the training and test set contain 132 trials, equally distributed for each class. The data was recorded at 1000 Hz using a multichannel system with 85 electrodes densely covering the motor cortex. After filtering, it was down-sampled to 100 Hz. The features are extracted as log-band power on CSP filtered channels (three filters per class) and Linear Discriminant Analysis (LDA) is used for classification.

The second set of recordings is the data set IVa \cite{DorBlaCurMul04} from BCI Competition III \cite{BlaMueKruSchWolSchPfuMilSchBir06} consisting of EEG recordings from five healthy subjects performing right hand and foot motor imagery without feedback. Two types of visual cues, a letters appearing behind a fixation cross and a randomly moving object, shown for 3.5s were used to indicate the target class. The presentation of target cues were intermitted by periods of random length, 1.75 to 2.25s, in which the subject could relax. The EEG signal was recorded from 118 Ag/AgCl electrodes, band-pass filtered between 0.05 and 200 Hz and downsampled to 100 Hz, so that 280 trials are available for each subject. We manually selected 68 electrodes densely covering the motor cortex and divided the data into a training and testing set based on the type of cue. Note that this division does not coincide with the one used for the competition, but in our experiments subjects B1 and B3 have 210 training trials (3 runs) and 70 test trials (1 run) and the other users have an equal number of 140 trials (2 runs) in each set. We extracted a time segment located from 500ms to 2500ms after the cue instructing the subject to perform motor imagery and band-pass filtered the signal in 8-30 Hz using a 5-th order Butterworth filter.

In addition to standard CSP we compute spatial filters with covCSP using the training covariance matrices of other subjects as regularization target and a wide range of trade-off parameters $\lambda = 0, 10^{-5}, 10^{-4}, 10^{-3}, 10^{-2}, 10^{-1}, .2, .3, .4, .5, .6, .7, .8, .9, 1$.
We also apply mtCSP using training data from other subjects and different trade-off parameters for $\lambda_1$ and $\lambda_2$, namely $10^{-4}, 10^{-3} \dots 10^{3}, 10^{4}$. The optimization is initialized with the spatial filters obtained by CSP. Finally the ssCSP approach is used with $l = 1 \ldots 8$ and $\nu = 1 \ldots 10$. We apply the same parameter selection scheme for all methods, namely we perform cross-validation in a leave-one-subject-out manner on the other subjects (using their training and test data sets) and use classification performance as selection criterion. In order to allow better comparison between methods and reduce complexity we do not use subject selection or subject clustering. Note that all analysis and interpretation is performed on the first data set.

\subsection{Initial Analysis}
In an initial analysis we study the similarity between users in order to evaluate whether multi-subject CSP methods are at all applicable. For this we first measure the distance between the covariance matrices of subjects $i$ and $j$ by symmetric Kullback-Leibler Divergence $\tilde{\mathrm{D}}_{KL} = \mathrm{D}_{KL}\left( \mathcal{N}(\mathbf{0},\matSigma_i)\ ||\ \mathcal{N}(\mathbf{0},\matSigma_j)\right) + \mathrm{D}_{KL}\left( \mathcal{N}(\mathbf{0},\matSigma_j)\ ||\ \mathcal{N}(\mathbf{0},\matSigma_i)\right)$\footnote{The Kullback-Leibler Divergence between Gaussians is defined as $D_\text{KL}(\mathcal{N}_0 \| \mathcal{N}_1) = { 1 \over 2 } \left( \mathrm{tr} \left( \Sigma_1^{-1} \Sigma_0 \right) + \left( \mu_1 - \mu_0\right)^\top \Sigma_1^{-1} ( \mu_1 - \mu_0 ) -\ln \left( { \det \Sigma_0 \over \det \Sigma_1  } \right) - k  \right).$}.
Table \ref{tab:kl} summarizes the results for each subject, it shows the average distance between the training/test covariance matrices of different subjects and the distance between training and test covariance matrix for the same user. One can see that variations between subjects are up to two orders larger than differences between training and test sessions. This indicates that transferring discriminative information between users may be highly unreliable. The divergence between training and test data is especially large in subject A4 and it is smallest in subject A5. These subjects also represent the two extreme cases in terms of classification accuracy (see Table \ref{tab:res}) which may indicate a correlation between the degree of stationarity and performance. However, since we do not test for significance, it may also be pure chance.

\begin{table*}
\begin{center}
\footnotesize
\caption{This table shows the average distance, measured by symmetric Kullback-Leibler Divergence, between the covariance matrices of different subjects (first and second row) and between the training and test covariance matrices for the same subject. We clearly see that the differences between subjects are up to two orders larger than the differences between training and test.}
\vspace*{0.2cm}
\hspace*{-1cm}
\onehalfspacing
\begin{tabular}{|l|p{0.5cm}p{0.5cm}p{0.5cm}p{0.5cm}c|}
\hline
Description & A1 & A2 & A3 & A4 & A5\\
\hline
Average $\tilde{\mathrm{D}}_{KL}$ to the training covariance matrices of other subjects& 490 & 799 & 650 & 853 & 657\\
Average $\tilde{\mathrm{D}}_{KL}$ to the test covariance matrices of other subjects& 995 & 1803 & 1799 & 1947 & 1377\\
$\tilde{\mathrm{D}}_{KL}$ between training and test covariance matrix for particular subject& 62 & 27 & 57 & 110 & 15\\
\hline
\end{tabular}
\label{tab:kl}
\end{center}
\end{table*}

\begin{figure}[h]
\begin{center}
\includegraphics[height=145px]{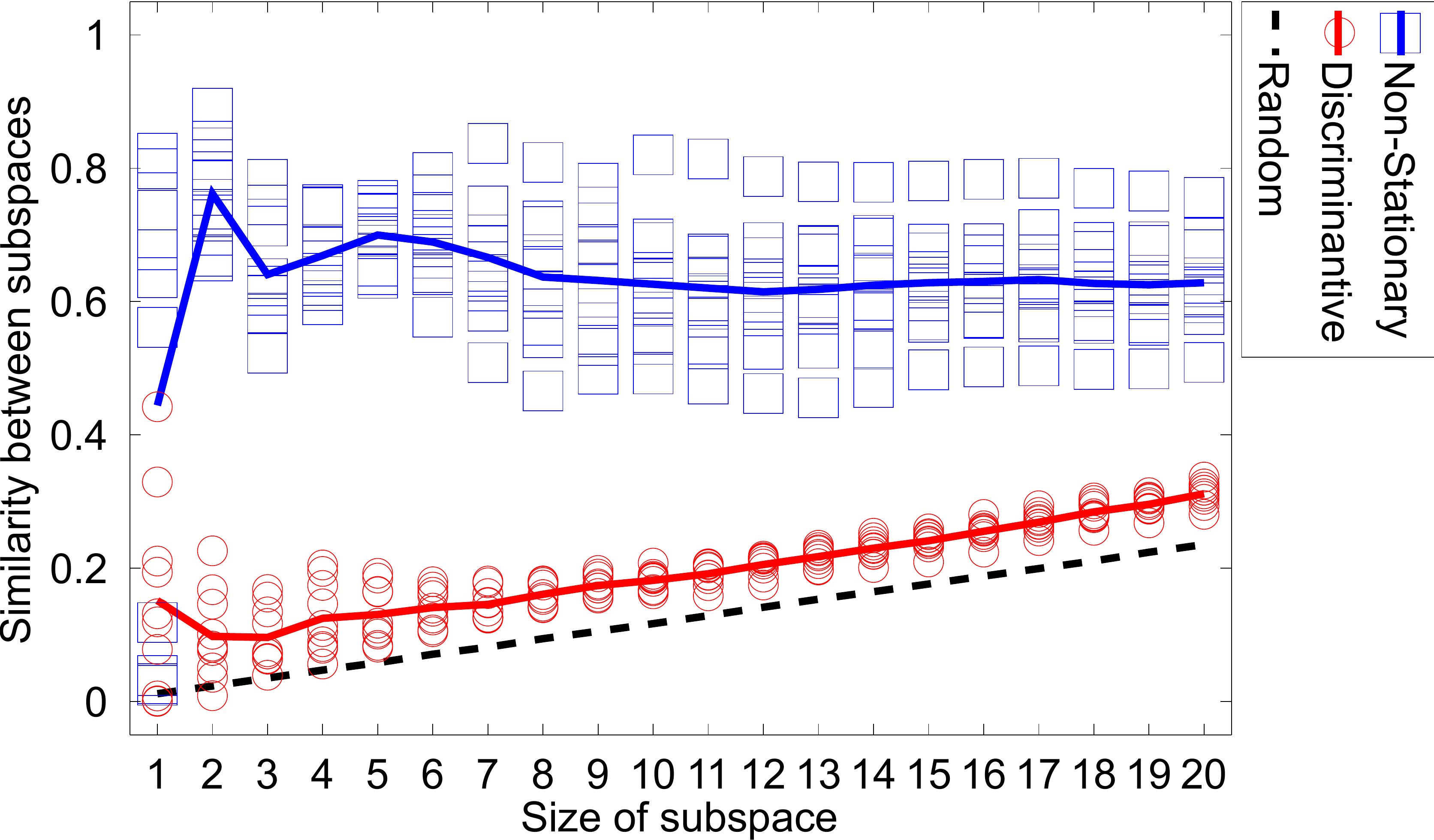}
\end{center}
\caption{Similarity between subspaces of different subjects measured as canonical correlation, or equivalently the mean of squared cosines of the principal angles. Each square and circle correspond to one comparison between two users, whereas the solid lines represent the mean similarities. We see that in contrast to the dominant non-stationary directions (blue line) the discriminant subspaces (red line) are quite different between subjects.}
\label{fig:angles}
\end{figure}

In Fig.~\ref{fig:angles} we analyse the similarity of subspaces extracted from different users.
We measure similarity as mean of squared cosines of the principal angles $\theta_k$ between the subspaces\footnote{Principal angles are defined recursively as $\cos(\theta_k) = \max_{u \in \mathcal{F}}\max_{v \in \mathcal{G}} u^Tv = u^T_kv_k$ subject to $||u|| = ||v|| = 1, \quad u^Tu_i = 0, \quad v^Tv_i = 0, \quad i = 1, \ldots, k - 1$. Note that there exist an equality between the canonical correlation and the cosine of principal angles.}. This corresponds to the amount of energy preserved when projecting data from one subspace to the other, thus higher values indicate closer subspaces. Considering all principal angles gives a clearer picture of the relation between two subspaces than when restricting the analysis to the largest principal angles as the latter one tends to become $90^\circ$ very fast. 
We extract two types of subspaces, namely discriminative and non-stationary ones. The discriminative subspace is constructed from the CSP spatial filters with largest eigenvalues. The non-stationary subspace is constructed from the prominent non-stationary directions (eigenvectors with largest absolute eigenvalues) between training and test.
From the plot we see that according to our measure of similarity the discriminative subspaces (red line) are not very similar between different users, the similarity is close to random (black dashed line), whereas the similarity between dominant non-stationary subspaces (blue line) is significant.
This is an important insight and the main motivation of our method. Note that we are not claiming that transferring discriminative information between subjects is impossible. Other measures of similarity exist that may better capture the amount of information contained in discriminative subspaces of other subjects, e.g.\ distances between class-conditional covariance matrices \cite{Lotte2010, Dev11}. The relation between those measures and the principle angles between subspaces is not trivial.

\subsection{Performance Comparison}
Table \ref{tab:res} summarizes the performance results for both data sets. We clearly see that performance can be improved by incorporating data from other users, however, not all subjects profit equally. As mentioned before ssCSP has a different focus than covCSP and mtCSP, namely it tackles the non-stationarity problem and not the estimation problem. Therefore it is not surprising that some users like A4, B1 and B3 significantly improve when mtCSP is applied and others like A1, A4 and B5 profit from the application of ssCSP. Note that the latter subjects have a large shift between training and test (see Table \ref{tab:kl}). We would also like to point out that in contrast to covCSP and mtCSP there is no significant decrease in performance when applying the ssCSP method. This observation is in line with the results from the toy experiment.
The bottom row of Table \ref{tab:res} shows the results of the combination of ssCSP and mtCSP with the regularization parameters obtained when applying both methods individually. In other words we first project out the non-stationary subspace obtained by ssCSP and then compute the spatial filters with mtCSP using the regularization parameters obtained when applying it to the original data. We see that this method gives the best performance results as it combines both transfer learning approaches.

We test the differences in performance statistically by applying a paired permutation test, i.e.\ we estimate an empirical distribution of mean performance differences using $2^{10}$ permutations (swapping the performances obtained with the different methods for each permutation of subjects) and compute the p-value for the actual difference. The p-values are summarized in Table \ref{tab:pval} and show that the improvement over the CSP baseline is significant up to 95\%.

\begin{table*}
\begin{center}
\footnotesize
\caption{Comparison of classification accuracies for different multi-subject CSP variants. All subjects profit from the information transfer except users B2. The best overall performance can be achieved by the combination of ssCSP and mtCSP.}
\vspace*{0.2cm}
\hspace*{-1cm}
\onehalfspacing
\begin{tabular}{|l|ccccc|ccccc||ccc|}
\hline
& \multicolumn{5}{c|}{Audio-Visual Data Set} & \multicolumn{5}{c||}{BCI Competition III} & \multicolumn{3}{c|}{Overall}\\
Subject & A1 & A2 & A3 & A4 & A5 & B1 & B2 & B3 & B4 & B5 & Mean & Median & Std\\
\hline
\hline
CSP & 79.5 & 80.0 & 65.8 & 59.2 & 94.2 & 66.1 & 96.4 & 58.2 & 88.8 & 81.0 & 76.9 & 79.8 & 14.0\\
covCSP & 78.8 & 75.0 & 61.7 & 60.8 & 95.0 & 71.4 & 96.4 & 70.4 & 73.7 & 89.7 & 77.3 & 74.3 & 12.7\\
mtCSP & 72.7 & 70.0 & 48.3 & 75.0 & 92.5 & 72.3 & 94.6 & 68.4 & 65.6 & 82.1 & 74.2 & 72.5 & 13.4\\
ssCSP & 87.1 & 80.8 & 67.5 & 65.8 & 93.3 & 67.0 & 94.6 & 58.2 & 89.3 & 85.7 & 78.9 & 83.3 & 13.1\\
ss+mtCSP & 87.9	& 80.8 & 66.7 & 69.2 & 93.3 & 71.4 & 94.6 & 66.3 & 88.4 & 84.9 & 80.4 & 82.9 & 11.1\\
\hline
\end{tabular}
\label{tab:res}
\end{center}
\end{table*}

\begin{table}
\begin{center}
\footnotesize
\caption{P-values computed by paired permutation test for the null hypothesis that there is no difference in mean performance between the methods.}
\onehalfspacing
\begin{tabular}{|l|cc|}
\hline
Method & ssCSP & ss+mtCSP\\
\hline
\hline
CSP & 0.0449 & 0.0224\\
covCSP & 0.2627 & 0.0820\\
mtCSP & 0.1191 & 0.0449\\
ssCSP & -- & 0.1094\\
\hline
\end{tabular}
\label{tab:pval}
\end{center}
\end{table}

\subsection{Interpretation}
In the following we analyse the non-stationarity activity patterns and investigate the reasons for the performance gain in more detail on the first subject A1.

\begin{figure}[h]
\begin{center}
\includegraphics[width=200px]{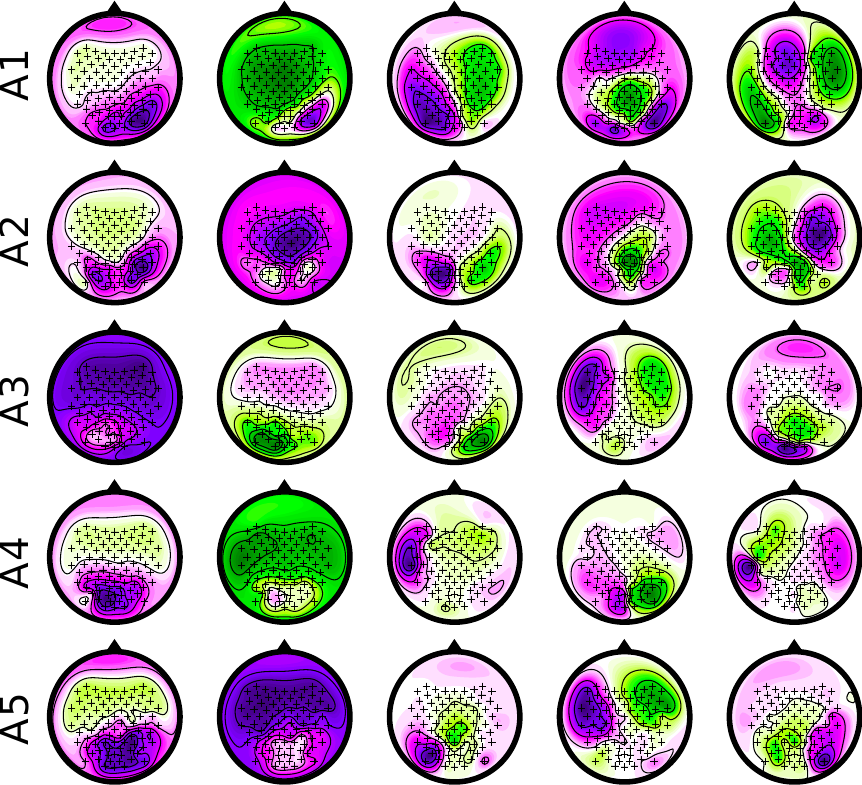}
\end{center}
\caption{Visualization of most non-stationary directions for each subject (in the rows). We clearly see that some of the patterns e.g.\ the first and third of subject A3, indicate a change in activity over occipital and temporal areas. These brain regions are mainly responsible for visual and auditory processing. Thus the principal non-stationary directions capture the change in the experimental conditions from a visual mode of stimulus presentation to an auditory one.}
\label{fig:patterns}
\end{figure}

Each row of Fig.~\ref{fig:patterns} visualizes the five most non-stationary directions of a subject. One can see that the patterns are highly similar between users. This similarity is also reflected in Fig.~\ref{fig:angles}. The non-stationarity patterns clearly show a relation to the change in the experimental conditions, i.e.\ the transition from a visual mode of stimulus presentation to an auditory one, as they focus mainly on occipital and temporal activity. From neuroscience it is well-known that occipital areas are responsible for visual processing and temporal regions are associated with auditory tasks. In other words the shift between training and test session is minimized by projecting out activity that is related to the presentation mode of the stimulus.

\begin{figure}[h]
\begin{center}
\includegraphics[height=162px]{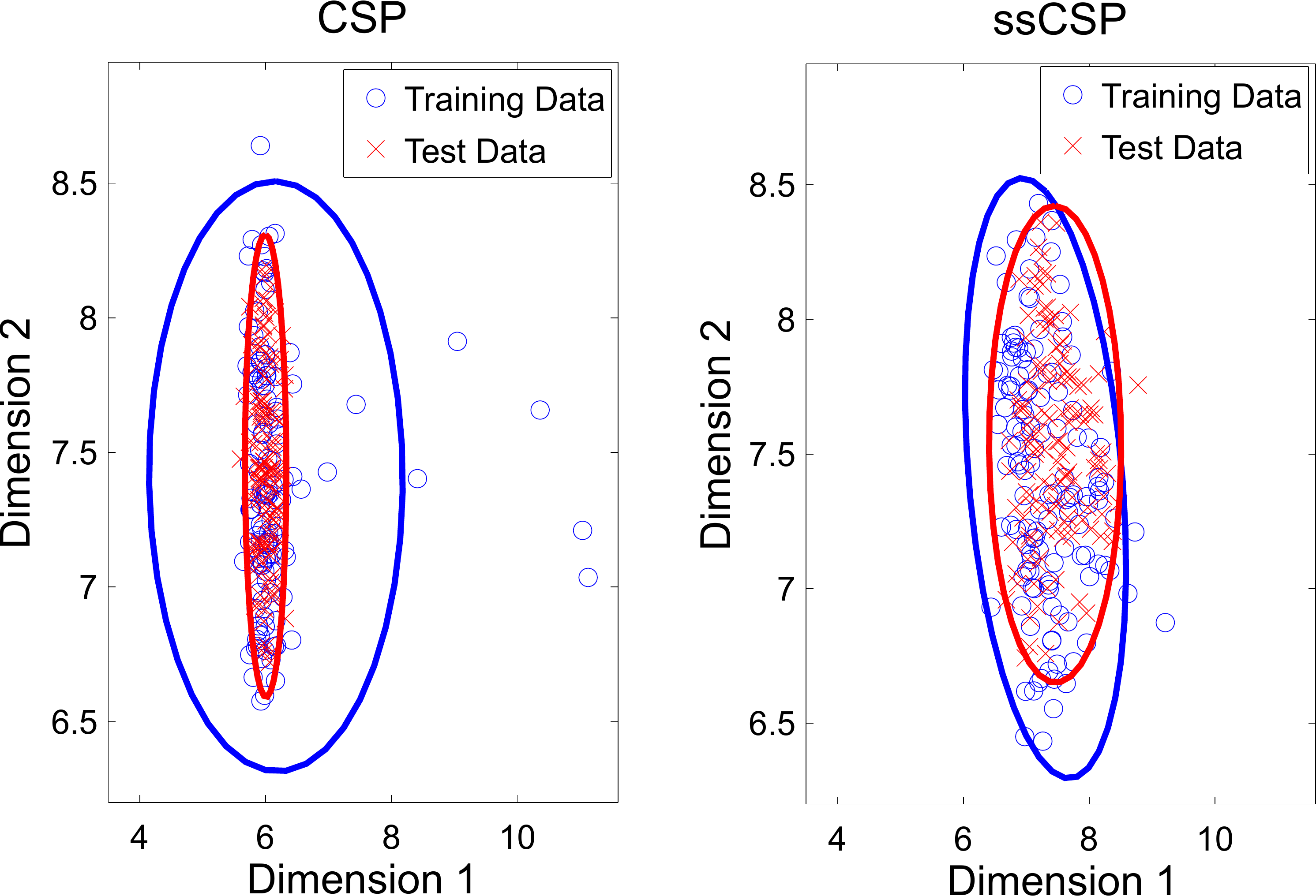}
\end{center}
\caption{Visualization of the two most discriminative dimensions for subject A1. A significant change in the feature distribution between training (blue circles) and test (red crosses) can be observed for the standard CSP method, whereas when applying ssCSP this change becomes almost negligible.}
\label{fig:shift}
\end{figure}

In Fig.~\ref{fig:shift} we see the change between the training and test features of subject 1 for CSP and ssCSP. We selected this user as he shows a significant increase in performance. We plot the two feature dimensions that correspond to the most discriminative filters in both conditions. We see that in the case of CSP the feature distribution obtained from training data is different from that computed on the test set. On the other hand when applying ssCSP there is only little difference between both distributions.

\subsection{Reducing Between-Day Variability}
In the previous subsections we showed that non-stationarities induced by changes in stimulation protocols may be transferred between subjects and used to extract invariant feature spaces. In this subsection we apply our transfer-learning approach to a different kind of variations, namely non-stationarities that occur when train- and test-sets have been recorded at different times. Reducing this between-day variability is crucial for zero-training BCI systems \cite{KraTanBlaMue08, fazli2009subject}.

The data set used for this experiment consists of recordings from five healthy subjects performing left and right hand motor imagery in five different calibration sessions. During the experiments the subjects were seated in a comfortable chair with arm rests and every $4.5-6$ seconds a visual stimuli was presented indicating the motor imagery task the subject should perform during the following $3-3.5$ seconds.
Between 140 and 288 trials were performed during one session and the sessions were recorded on different days. The data set contains recordings from 48 channels densely located over the central areas of the scalp. We apply a fixed preprocessing scheme for all subjects, i.e.\ we extract the $750-3500$ms time segment after the cue and band-pass filter the signal in $8-30$Hz. For each subject we use one session as train set and the other four sessions as test sets. The between-day variability and the parameters of ssCSP are estimated from other subjects in the same manner as before.

The mean classification accuracy of each subject when training on the first session and testing on the others is shown in Table \ref{tab:res2}. As in the previous experiment one can observe a performance increase when applying transfer learning, however, the effect is rather small. The main reason for the reduced improvement is a lower similarity score between the prominent non-stationarities of different subjects. This indicates that between-day variability is less stable across subjects than non-stationarities induced by differences in experimental conditions.

\begin{table}
\begin{center}
\footnotesize
\caption{Mean classification accuracies for the session-to-session transfer experiment.}
\onehalfspacing
\begin{tabular}{|l|ccccc|}
\hline
Method & Sub1 & Sub2 & Sub3 & Sub4 & Sub5\\
\hline
\hline
CSP & 71.5 & 52.8 & 62.0 & 92.2 & 62.6\\
ssCSP & 70.2 & 54.6 & 69.1 & 91.7 & 63.7\\
\hline
\end{tabular}
\label{tab:res2}
\end{center}
\end{table}

\subsection{Learning from Noise ?}
\begin{figure}[h]
\begin{center}
\includegraphics[height=120px]{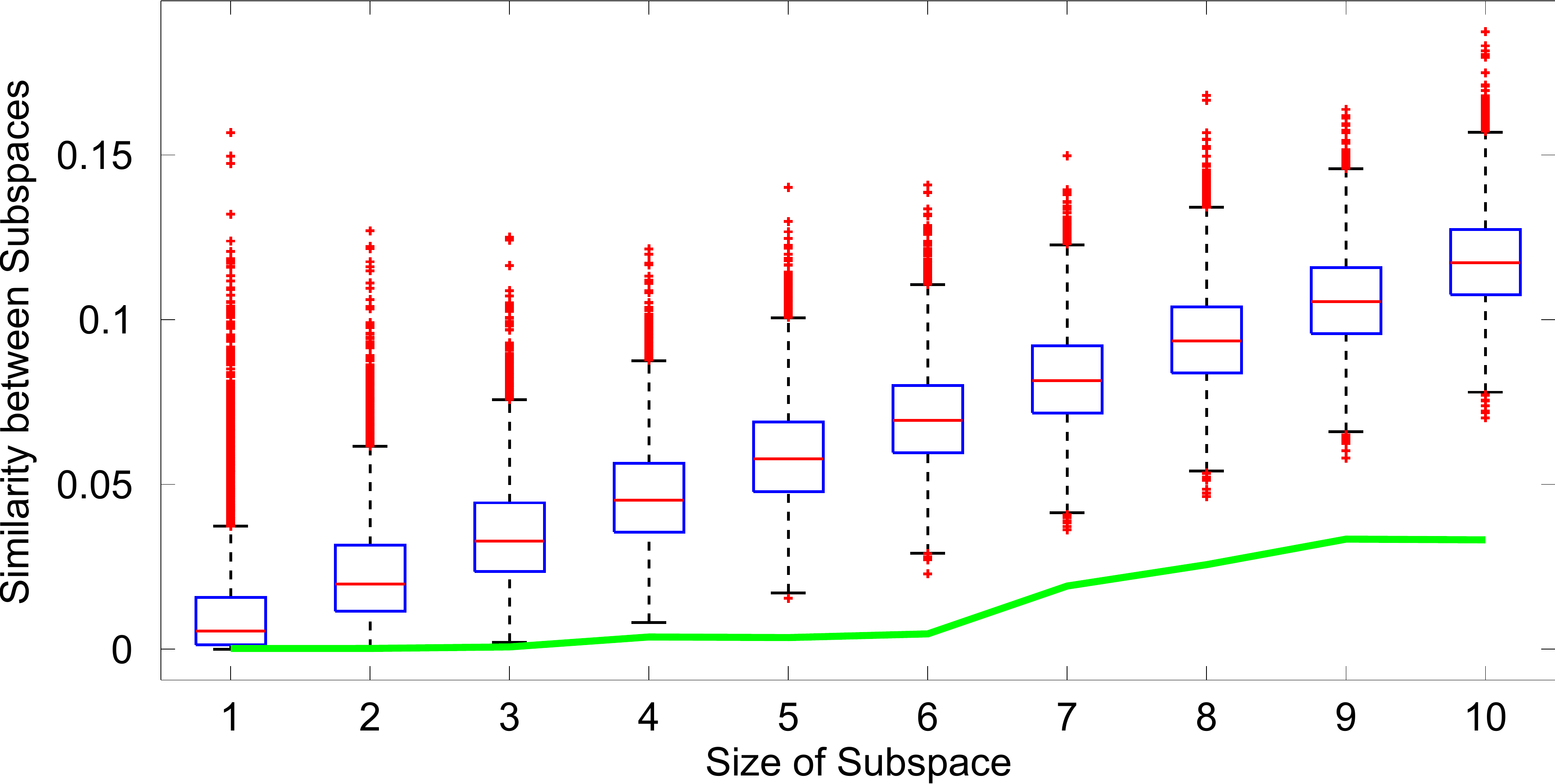}
\end{center}
\caption{Boxplot showing the empirical distribution of similarity scores between the CSP subspace and random subspaces for different dimensionality. The solid green line denotes the similarity between the CSP subspace and the non-stationary subspace of subject A5. One can see that the similarity between the discriminative and non-stationary subspaces is much smaller than between the discriminative subspace and a random one.}
\label{fig:rand}
\end{figure}

An interesting question is whether the prominent changes occur in the discriminative or in the non-discriminative part of the signal.
In other words we investigate the similarity between the subspaces spanned by the most non-stationary directions and the most discriminative ones. If the subspaces are dissimilar then most changes occur in the non-discriminative part of the signal.
In order to study this question we compute the similarity scores between the subspace spanned by CSP and the non-stationary subspaces (up to dimension 10) for each subject. As before we measure similarity as mean square cosine of principal angles. Additionally, we estimate the empirical distribution of these similarity scores for each dimensionality by comparing the CSP subspace to 10000 randomly generated subspaces. 
It turns out that the actual similarities all lie in the lower 1\% quantile of the corresponding empirical distribution (see Fig.\ \ref{fig:rand}). This indicates that the similarity between the discriminative and non-stationary subspaces is significantly smaller than random, consequently most of the shift is present in the non-discriminative part of the data.

In order to assess how relevant the shift in the non-discriminant subspace is, we project out the (discriminative) CSP directions from the data of each subject prior to computation of the non-stationary subspace.
When applying this approach to both data sets we obtain an average performance of 78.1 i.e.\ the performance loss compared to the original ssCSP method (78.9) is minimal and not significant. This is a surprising result as it indicates that the non-discriminative ´noise´ signal subspace can aid to construct invariant features. This subspace is generally removed (by applying CSP) prior to classification and regarded as non-task related noise.
Thus we need to revisit the statement that {\it noise never helps} as it can be used to improve classification accuracy and reduce the need of adaptation in a BCI scenario.

%%%%%%%%%%%%%%%%%%%%%%%%%%%%%%%%%%%%%%%%%%%%%%%%%%%%%%%%%%%%%%%%%%%%%%%%%%%%%%%%%%%
\section{Discussion}
Non-stationarities in BCI experiments are rather common and they are notoriously hard to model. In this work we showed that information about dominant changes can be transferred between subjects and is mainly contained in the non-discriminant (noise) part of the data. Thus, somewhat
paradoxically, the noise part can be the key to improve classification accuracy, as it allows to define invariant features.
We showed quantitatively that prominent non-stationarities resulting from changes in the experimental conditions are much more stably estimated between subjects than their respective discriminant (information carrying) subspaces. Note that the non-stationarity information transferred between subject appears physiologically interpretable and meaningful. Moreover reducing non-stationarities from data is seen to be more robust to perturbations than learning discriminative subspaces, thus subject selection or weighting is not required. We will in the future investigate theoretical limits and applications of our concept to transfer learning and covariate shift models. Finally we intend to evaluate our approach in an online BCI setting and investigate ways to transfer information obtained from different imaging modalities \cite{BiePliMeiEicMue11, fazli2011enhanced}.

%%%%%%%%%%%%%%%%%%%%%%%%%%%%%%%%%%%%%%%%%%%%%%%%%%%%%%%%%%%%%%%%%%%%%%%%%%%%%%%%%%%
\section*{Acknowledgment}
This work was supported by the German Research Foundation (GRK 1589/1), by the Federal Ministry of Education and Research (BMBF) under the project Adaptive BCI (FKZ 01GQ1115) and by the World Class University Program through the National Research Foundation of Korea funded by the Ministry of Education, Science, and Technology, under Grant R31-10008.

\ifCLASSOPTIONcaptionsoff
  \newpage
\fi

\bibliographystyle{IEEEtran}
\bibliography{bme2012}

\begin{thebibliography}{10}
\providecommand{\url}[1]{#1}
\csname url@samestyle\endcsname
\providecommand{\newblock}{\relax}
\providecommand{\bibinfo}[2]{#2}
\providecommand{\BIBentrySTDinterwordspacing}{\spaceskip=0pt\relax}
\providecommand{\BIBentryALTinterwordstretchfactor}{4}
\providecommand{\BIBentryALTinterwordspacing}{\spaceskip=\fontdimen2\font plus
\BIBentryALTinterwordstretchfactor\fontdimen3\font minus
  \fontdimen4\font\relax}
\providecommand{\BIBforeignlanguage}[2]{{%
\expandafter\ifx\csname l@#1\endcsname\relax
\typeout{** WARNING: IEEEtran.bst: No hyphenation pattern has been}%
\typeout{** loaded for the language `#1'. Using the pattern for}%
\typeout{** the default language instead.}%
\else
\language=\csname l@#1\endcsname
\fi
#2}}
\providecommand{\BIBdecl}{\relax}
\BIBdecl

\bibitem{KraTanBlaMue08}
M.~Krauledat, M.~Tangermann, B.~Blankertz, and K.-R. M\"uller, ``Towards zero
  training for brain-computer interfacing,'' \emph{PloS one}, vol.~3, no.~8, p.
  e2967, 2008.

\bibitem{Dev11}
D.~Devlaminck, B.~Wyns, M.~Grosse-Wentrup, G.~Otte, and P.~Santens,
  ``Multi-subject learning for common spatial patterns in motor-imagery bci,''
  \emph{Computational Intelligence and Neuroscience}, vol. 2011, no. 217987,
  pp. 1--9, 2011.

\bibitem{Ala10}
M.~Alamgir, M.~Grosse-Wentrup, and Y.~Altun, ``Multitask learning for
  brain-computer interfaces,'' in \emph{JMLR Workshop and Conference
  Proceedings Volume 9: AISTATS 2010, Thirteenth International Conference on
  Artificial Intelligence and Statistics}, 2010, pp. 17--24.

\bibitem{Lotte2010}
F.~Lotte and C.~Guan, ``Learning from other subjects helps reducing
  {Brain-Computer} interface calibration time,'' in \emph{ICASSP'10: 35th IEEE
  International Conference on Acoustics, Speech, and Signal Processing}, 2010,
  pp. 614--617.

\bibitem{kan09}
H.~Kang, Y.~Nam, and S.~Choi, ``Composite common spatial pattern for
  subject-to-subject transfer,'' \emph{Signal Processing Letters, IEEE},
  vol.~16, no.~8, pp. 683 --686, 2009.

\bibitem{Can09}
J.~Quionero-Candela, M.~Sugiyama, A.~Schwaighofer, and N.~D. Lawrence,
  \emph{Dataset Shift in Machine Learning}.\hskip 1em plus 0.5em minus
  0.4em\relax The MIT Press, 2009.

\bibitem{SugKaw2011}
M.~Sugiyama and M.~Kawanabe, \emph{Machine learning in non-stationary
  environments: Introduction to covariate shift adaptation}.\hskip 1em plus
  0.5em minus 0.4em\relax Cambridge, MA: MIT Press, 2011.

\bibitem{SheKraBlaRaoMue06}
P.~Shenoy, M.~Krauledat, B.~Blankertz, R.~P. Rao, and K.-R. M\"{u}ller,
  ``Towards adaptive classification for {BCI},'' \emph{Journal of neural
  engineering}, vol.~3, no.~1, pp. R13--R23, 2006.

\bibitem{SugKraMue07}
M.~Sugiyama, M.~Krauledat, and K.-R. M\"{u}ller, ``Covariate shift adaptation
  by importance weighted cross validation,'' \emph{J. Mach. Learn. Res.},
  vol.~8, pp. 985--1005, 2007.

\bibitem{reuderink2011robust}
B.~Reuderink, ``Robust brain-computer interfaces,'' Ph.D. dissertation,
  University of Twente, 2011.

\bibitem{Blankertz08optimizingspatial}
B.~Blankertz, R.~Tomioka, S.~Lemm, M.~Kawanabe, and K.-R. M\"{u}ller,
  ``{Optimizing Spatial filters for Robust EEG Single-Trial Analysis},''
  \emph{IEEE Signal Proc. Magazine}, vol.~25, no.~1, pp. 41--56, 2008.

\bibitem{Ramoser98optimalspatial}
H.~Ramoser, J.~M\"{u}ller-Gerking, and G.~Pfurtscheller, ``Optimal spatial
  filtering of single trial eeg during imagined hand movement,'' \emph{IEEE
  Trans. Rehab. Eng.}, vol.~8, no.~4, pp. 441--446, 1998.

\bibitem{LemmBlaDicMue11}
S.~Lemm, B.~Blankertz, T.~Dickhaus, and K.-R. M\"{u}ller, ``Introduction to
  machine learning for brain imaging,'' \emph{NeuroImage}, vol.~56, no.~2, pp.
  387--399, 2011.

\bibitem{LotGua11}
F.~Lotte and C.~Guan, ``Regularizing common spatial patterns to improve bci
  designs: Unified theory and new algorithms,'' \emph{IEEE Trans. Biomed.
  Eng.}, vol.~58, no.~2, pp. 355 --362, 2011.

\bibitem{SamJNE12}
W.~Samek, C.~Vidaurre, K.-R. M{\"u}ller, and M.~Kawanabe, ``Stationary common
  spatial patterns for brain-computer interfacing,'' \emph{Journal of Neural
  Engineering}, vol.~9, no.~2, p. 026013, 2012.

\bibitem{Arv13}
M.~Arvaneh, C.~Guan, K.~K. Ang, and C.~Quek, ``Optimizing spatial filters by
  minimizing within-class dissimilarities in electroencephalogram-based
  brain-computer interface,'' \emph{Neural Networks and Learning Systems, IEEE
  Transactions on}, vol.~24, no.~4, pp. 610--619, April.

\bibitem{Blankertz08invariantcommon}
B.~Blankertz, M.~K.~R. Tomioka, F.~U. Hohlefeld, V.~Nikulin, and K.-R.
  M\"{u}ller, ``Invariant common spatial patterns: Alleviating
  nonstationarities in brain-computer interfacing,'' in \emph{Ad. in NIPS 20},
  2008, pp. 113--120.

\bibitem{BueMeiKirMue09}
P.~von B\"unau, F.~C. Meinecke, F.~C. Kir\'aly, and K.-R. M\"uller, ``Finding
  stationary subspaces in multivariate time series,'' \emph{Phys. Rev. Lett.},
  vol. 103, p. 214101, Nov 2009.

\bibitem{Sam12}
W.~Samek, K.-R. M\"uller, M.~Kawanabe, and C.~Vidaurre, ``Brain-computer
  interfacing in discriminative and stationary subspaces,'' in \emph{IEEE Int.
  Conf. of Engineering in Medicine and Biology Society (EMBC)}, 2012.

\bibitem{Kra08}
M.~Krauledat, ``Analysis of nonstationarities in eeg signals for improving
  brain-computer interface performance,'' Ph.D. dissertation, Technische
  Universit{\"a}t Berlin, 2008.

\bibitem{fazli2009subject}
S.~Fazli, F.~Popescu, M.~Dan{\'o}czy, B.~Blankertz, K.-R. M{\"u}ller, and
  C.~Grozea, ``Subject-independent mental state classification in single
  trials,'' \emph{Neural networks}, vol.~22, no.~9, pp. 1305--1312, 2009.

\bibitem{VidSanMueBla10}
C.~Vidaurre, C.~Sannelli, K.-R. M\"{u}ller, and B.~Blankertz,
  ``Machine-learning-based coadaptive calibration for brain-computer
  interfaces,'' \emph{Neural Comp.}, vol.~23, no.~3, pp. 791--816, 2011.

\bibitem{VidSanMueBla11}
C.~Vidaurre, C.~Sannelli, K.-R. M{\"u}ller, and B.~Blankertz,
  ``Machine-learning-based coadaptive calibration for brain-computer
  interfaces,'' \emph{Neural Computation}, vol.~23, no.~3, pp. 791--816, 2011.

\bibitem{DorBlaCurMul04}
G.~Dornhege, B.~Blankertz, G.~Curio, and K.-R. M\"{u}ller, ``Boosting bit rates
  in noninvasive eeg single-trial classifications by feature combination and
  multiclass paradigms,'' \emph{IEEE Trans. Biomed. Eng.}, vol.~51, no.~6, pp.
  993 --1002, 2004.

\bibitem{BlaMueKruSchWolSchPfuMilSchBir06}
B.~Blankertz, K.-R. M\"{u}ller, D.~Krusienski, G.~Schalk, J.~Wolpaw,
  A.~Schl\"ogl, G.~Pfurtscheller, J.~del R.~Mill\'an, M.~Schr\"oder, and
  N.~Birbaumer, ``The bci competition iii: validating alternative approaches to
  actual bci problems,'' \emph{IEEE Trans. on Neural Syst. and Rehabil. Eng.},
  vol.~14, no.~2, pp. 153 --159, 2006.

\bibitem{BiePliMeiEicMue11}
F.~Bie{\ss}mann, S.~M. Plis, F.~C. Meinecke, T.~Eichele, and K.-R. M{\"u}ller,
  ``Analysis of multimodal neuroimaging data,'' \emph{IEEE Rev. Biomed. Eng.},
  vol.~4, pp. 26 -- 58, 2011.

\bibitem{fazli2011enhanced}
S.~Fazli, J.~Mehnert, J.~Steinbrink, G.~Curio, A.~Villringer, K.-R. M\"uller,
  and B.~Blankertz, ``Enhanced performance by a hybrid nirs–eeg brain
  computer interface,'' \emph{NeuroImage}, vol.~59, no.~1, pp. 519 -- 529,
  2012.

\end{thebibliography}
\end{document}